\documentclass{article}

    \usepackage[preprint, nonatbib]{neurips_2023}

\usepackage[utf8]{inputenc} 
\usepackage[T1]{fontenc}    
\usepackage{hyperref}       
\usepackage{url}            
\usepackage{booktabs}       
\usepackage{amsfonts}       
\usepackage{nicefrac}       
\usepackage{microtype}      
\usepackage{xcolor}         
\usepackage{graphicx}
\usepackage{amsmath}
\usepackage{comment}
\usepackage{wrapfig}

\DeclareMathOperator*{\argmin}{arg\,min}

\title{Linear Mode Connectivity in Sparse Neural Networks}

\author{
  Luke McDermott \\
  Modern Intelligence\\
  \texttt{luke@modernintelligence.ai}\\
  \And
  Daniel Cummings \\
  Modern Intelligence\\
  \texttt{daniel@modernintelligence.ai}
}

\begin{document}

\maketitle

\begin{abstract}
With the rise in interest of sparse neural networks, we study how neural network pruning with synthetic data leads to sparse networks with unique training properties. We find that distilled data, a synthetic summarization of the real data, paired with Iterative Magnitude Pruning (IMP) unveils a new class of sparse networks that are more stable to SGD noise on the real data, than either the dense model, or subnetworks found with real data in IMP. That is, synthetically chosen subnetworks often train to the same minima, or exhibit linear mode connectivity. We study this through linear interpolation, loss landscape visualizations, and measuring the diagonal of the hessian. While dataset distillation as a field is still young, we find that these properties lead to synthetic subnetworks matching the performance of traditional IMP with up to 150x less training points in settings where distilled data applies.
\end{abstract}

\section{Introduction \& Background}
Sparse neural networks are increasingly important in deep learning to enhance hardware performance (e.g., memory footprint, inference time) and reduce environmental impacts (e.g., energy consumption), especially as state-of-the-art foundational models continue to grow significantly in parameter count. The most common form of sparsity can be found in neural network pruning literature \cite{PruneSurvey}. In this field, researchers exploit sparsity for computational savings, usually at inference, by removing parameters after training. In order to reduce the cost of training as well, other works explore how to prune at initialization \cite{SNIP, SynFlow}, the end goal for almost any pruning research. Despite these great ambitions, pruning at initialization does not perform as we hope \cite{PruneInitWhyMissing}. To further understand why this is the case, Frankle et al. \cite{LTH} propose the Lottery Ticket Hypothesis: \textit{for a sufficiently over-parameterized dense network, there exists a non-trivial sparse subnetwork that can train in isolation to the full performance of the dense model}. This is empirically validated for small settings with Iterative Magnitude Pruning with weight rewinding back to initialization. In parallel, researchers have been exploring how synthetic data representations such as those generated by dataset distillation methods can be leveraged to efficiently accelerate deep learning model training. With this in mind, we explore the training dynamics and stability of sparse neural networks in the context of synthetic data to better understand how we should be efficiently creating sparsity masks at initialization. 

Research on the training dynamics of dense models have led researchers to find that dense models are connected in the loss landscape through nonlinear paths \cite{NonlinearPath1, NonlinearPath2, NonlinearPath3}. Linear paths or Linear Mode Connectivity (LMC) is an uncommon phenomena that only occurs in rare cases, such as MLPs on subsets of MNIST in \cite{FirstLMC}. For large networks, Frankle et al. \cite{LMCLTH} found that pretrained dense models, when fine-tuned across different shufflings of data, are linear mode connected. While these models are “stable” to noise generated through stochastic gradient descent (SGD), only the smallest dense models are stable at initialization. 
As for its relationship with sparse neural networks, it was empirically found that the Lottery Ticket Hypothesis only holds for stable dense models, those that are linear mode connected across data shuffling \cite{LMCLTH}. They found that these large dense models only become stable early in training, leading to the conclusion that Iterative Magnitude Pruning (IMP) with weight rewinding, the method to find such lottery tickets, should instead rewind a model to an early point in training rather than at initialization, revising the hypothesis to fit in larger settings. Our work aims to study the properties of sparse neural networks at initialization, different than new age lottery ticket literature \cite{DataDiet}, which utilizes some pretraining to find a "good" initialization. 

We find that another class of sparse subnetworks exist that are more stable at initialization: \textit{synthetic subnetworks}. We define synthetic subnetworks as those produced during "distilled pruning" \cite{DistilledPruning}. These are found by replacing the traditional data in IMP with distilled data, essentially a summarized version of the training data consisting of only 1-50 synthetic images per class (see \cite{DatasetDistillationSurvey} for a survey). In general, dataset distillation optimizes a synthetic dataset to match the performance of a model trained on real data. This bi-level optimization problem can be defined as minimizing the difference of average loss over all validation points:

\begin{equation}
    \argmin_{\mathcal{D}_{\text{syn}}} | L(\Phi(\mathcal{D}_{\text{real}}); \mathcal{D}_{\text{val}}) - L(\Phi(\mathcal{D}_{\text{syn}}); \mathcal{D}_{\text{val}}) | 
\end{equation}

In distilled pruning, we perform the same training, pruning, and rewinding to initialization in order to produce the sparsity mask. This mask, as with those produced by IMP, can be applied to the dense model at initialization to create a high performing sparse neural network after training on real data. The significance is that synthetic images can be used to pick an appropriate sparsity mask for a downstream task. Recent work shows that despite synthetic subnetworks having a lower performance as a trade-off for pruning efficiency, these subnetworks have a lower need for rewinding to an early point in training due to their inherent stability \cite{DistilledPruning}. We find that, with better dataset distillation methods such as Information-intensive Dataset Condensation (IDC) \cite{EfficientDatasetCondensation} rather than Matching Training Trajectories (MTT) \cite{MTT} which was used previously, we match performance with IMP when rewinding to initialization. We achieve this with 5x less data than previous distilled pruning work which is approximately \textit{150x less} training points than traditional IMP to find a sparsity mask. While we do use a current state-of-the-art distillation method, such methods are still limited to models up to ResNet-18 and small datasets like CIFAR-10, CIFAR-100, and subsets of ImageNet.

\section{Dataset distillation for neural network pruning.} 
To find a suitable sparsity mask of a randomly initialized model, we first train the network to convergence on distilled data\footnote{We use dataset distillation methods that match training trajectories to ensure that training on synthetic data yields similar converged results to training on real data \cite{EfficientDatasetCondensation}}, prune the lowest magnitude weights, then rewind the non-pruned weights back to their initialized values, and loop until desired sparsity. The final model should have its randomly initialize weights with a sparsity mask. We can train the sparse synthetic subnetwork on real data to achieve sufficient performance at high sparsities. Using distilled data only to choose our sparsity mask allows us to better understand the architectural relationship of this data. We refer to subnetworks found with synthetic or distilled data as \textit{synthetic subnetworks} and those with real data as \textit{IMP subnetworks}. The only differences of IMP and distilled pruning lie in the sparsity mask they choose. Since each method uses different datasets for training, their final converged weights will be different. What is deemed "important" for real data might not be important for distilled data; therefore, distilled pruning may attempt to remove these. The performance of these sparsity masks by distilled pruning directly relates to how relevant the distilled data is to the real data. We find that distilled pruning can match the performance of IMP, when rewinding to initialization, on settings where dataset distillation applies\footnote{In the appendix, figure \ref{fig:SVATTM} showcases this performance, comparing traditional IMP to distilled pruning on CIFAR-10 with ResNet-18. }.

\section{Stability of subnetworks.}
\begin{figure}[htb]
  \centering
  \includegraphics[width=5in]{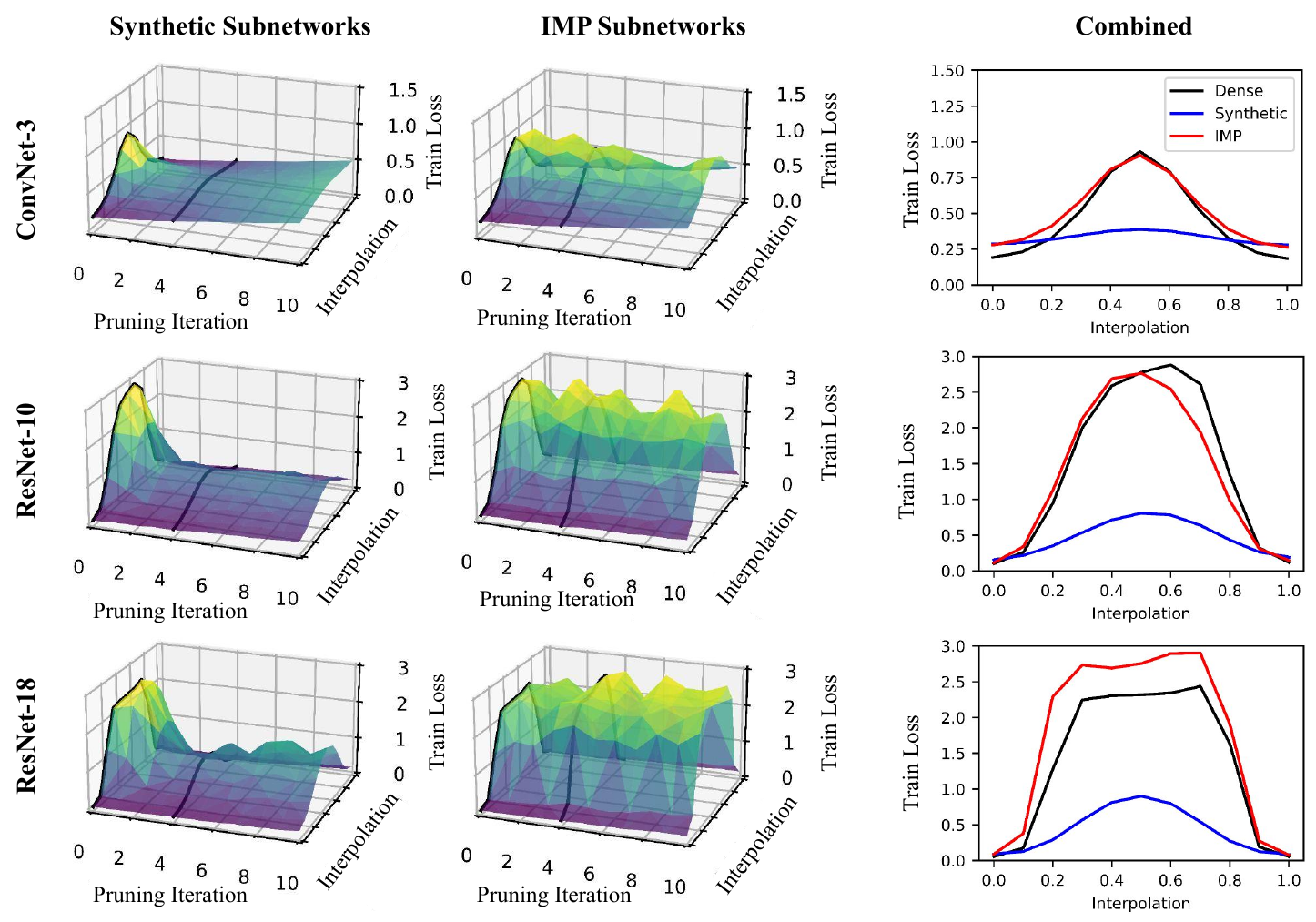}
  \caption{Comparison of the stability of synthetic vs. IMP subnetworks at initialization on CIFAR-10. We show how the loss increases as you interpolate the weights between two trained models. We measure this for subnetworks of different sparsities. The left column is reserved for subnetworks found via distilled data, and the middle column is for subnetworks found with real data. 
  The dark lines in the 3D plots represents the pruning iteration we used for the combined plot; the dense model is iteration 0. }
  \label{fig:LMC_cifar10}
\end{figure}

\begin{figure}[htb]
  \centering
  \makebox[0pt]{\includegraphics[width=12cm]{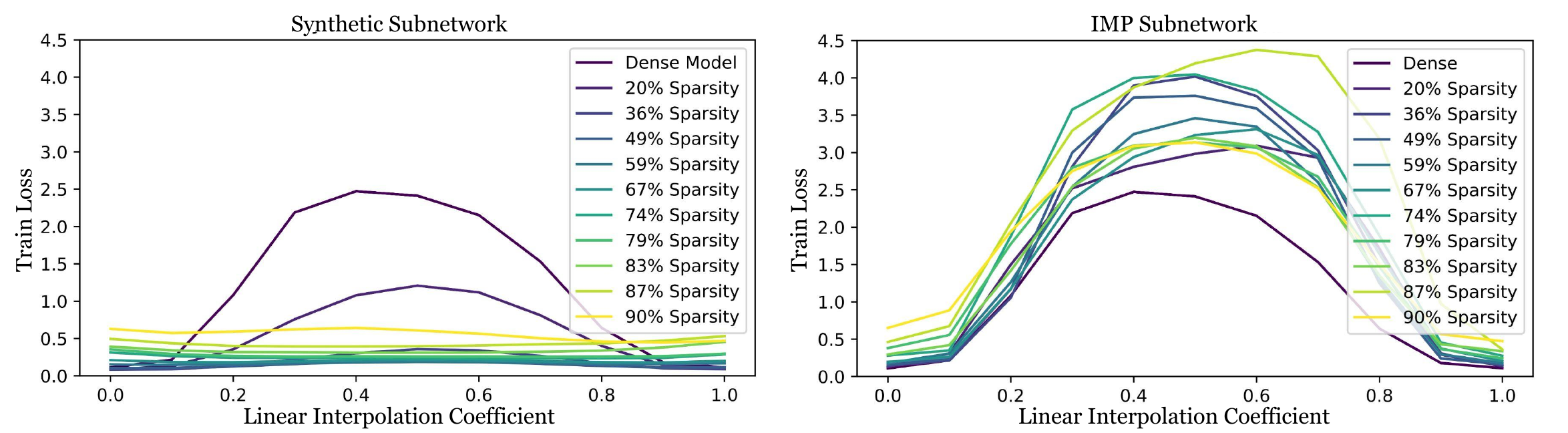}}
  \caption{Comparison of the stability of synthetic vs. IMP subnetworks at initialization on ImageNet-10 and ResNet-10. An increased loss across interpolation implies instability / trained networks landing in different minima. }
  \label{fig:LMC_imagenet10}
\end{figure}

To understand the training dynamics of sparsity masks chosen via distilled pruning vs IMP, we conduct an instability analysis. We take a randomly initialized model, generate a sparsity mask through pruning, and train it across two different orderings of the real training data. We save these two models and interpolate all the weights between them, measuring the training loss at each point in the interpolation as shown in Figures \ref{fig:LMC_cifar10} and \ref{fig:LMC_imagenet10}. We assess the linear mode connectivity of these subnetworks to determine if the model is stable to SGD noise. If the loss increases as you interpolate between two trained versions, then there is a barrier in the loss landscape, implying the trained models found different minima. In these cases, the ordering of the training data directly impacts what minima its choosing. If the loss does not increase during interpolation, then this implies they exist in the same minima or at least the same flat basin. 

We see that in simpler scenarios with ConvNet-3 on CIFAR-10 \& ResNet-10 on ImageNet-10, we exhibit full linear mode connectivity. We even see slightly better performance during interpolation in Figure \ref{fig:LMC_imagenet10}. As stated before, it was shown that lottery tickets can be found with IMP only when the dense model is stable \cite{LMCLTH}. However, we find that in some cases of unstable dense models there exists a sparse subnetwork that is stable at initialization. More importantly, traditional IMP is not able to produce stable subnetworks in these settings. Sparsity is not necessarily the answer for smoother landscapes, \textit{where} you induce sparsity is the main factor.  As pruning continues, the results exhibit more stability despite lower trainability, as seen with higher training losses. We postulate that the parameters pruned on distilled data, yet still exist in the IMP subnetwork, capture the intricacies of the real data which contribute to a sharper, but more trainable, landscape. Since IMP subnetworks are not stable, the intricacies it is learning is order dependent.


\section{Loss Landscape Visualization}
\begin{figure}[htb]
\centering
  \includegraphics[width=11cm]{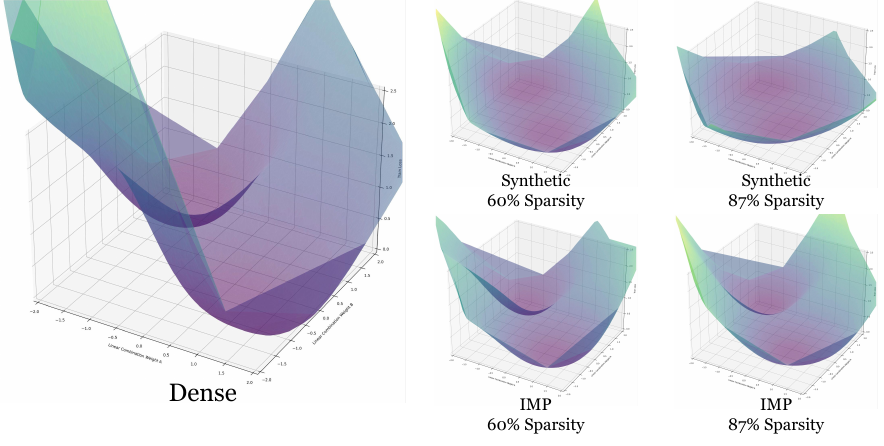}
  \caption{Loss Landscape visualization around the neigbhorhood defined by trained models on different seeds for ConvNet-3 and CIFAR-10.}
  \label{fig:loss_landscape}
\end{figure}

While linear mode connectivity is useful to study the loss landscape, this lightweight method can only show us a one dimensional slice of the bigger picture. We further examine the landscapes across two dimensions of parameters as shown in Figure \ref{fig:loss_landscape}. 

We created two orthogonal vectors from trained reference models in order to map the hyperdimensional parameter space down to two dimensions. For each of the 10,000 points, we take the linear combination of the two vectors, and measure loss on real training data. Since this visualization is created after reference models are trained, reference models that are closer together will result in “zooming in” on their minima. The spatial distance is not preserved using this method. This is useful in determining the local area in which these models are training to. With post-hoc analysis, we find that spatial distance in our plot is mainly maintained, with slightly lower distances as you prune. From these visualizations, IMP chooses subnetworks that exhibit a similar landscape to the dense model. We see the trained models fall into two separate minima in both the IMP and Dense cases, explaining the loss barrier in the Figure \ref{fig:LMC_cifar10}. Subnetworks chosen with distilled data are falling into the same, flat basin.  

Across almost all experiments, we see a general trend: subnetworks chosen via distilled pruning result in a smooth \& generalizing loss landscape. As compression ratio increases, we see more stability than IMP; however, the performance trend largely depends on the distilled accuracy, in this case by using the IDC method \cite{EfficientDatasetCondensation}. Most notably, we achieve full linear mode connectivity for ConvNet-3 on CIFAR-10 and ResNet-10 on Imagenet-10. While there are numerous factors at play, IDC \cite{EfficientDatasetCondensation} optimized the synthetic data specifically for these models on each dataset, hinting that stability is a result of high performing synthetic data. 

\section{Conclusion}
This work is an initial step into exploring the impact of using synthetic data, specifically distilled data, on pruning. We thoroughly assess the linear mode connectivity of these subnetworks to determine if the model is stable to SGD noise, even finding stable subnetworks from unstable dense models. We believe the inherent compression of dataset distillation is a driving factor in synthetic subnetworks' stability. Lastly, we believe this hints at the possibility of finding lottery tickets at initialization by first searching for stable subnetworks. In turn, we invite researchers to find new ways to search for stable subnetworks, especially on the real data.

\newpage
\small
\bibliographystyle{plain}
\bibliography{bibliography.bib}

\normalsize

\newpage
\appendix
\section{Synthetic Data \& The Information Bottleneck}
\begin{figure}[htb]
  \centering
  \makebox[0pt]{\includegraphics[width=5.5in]{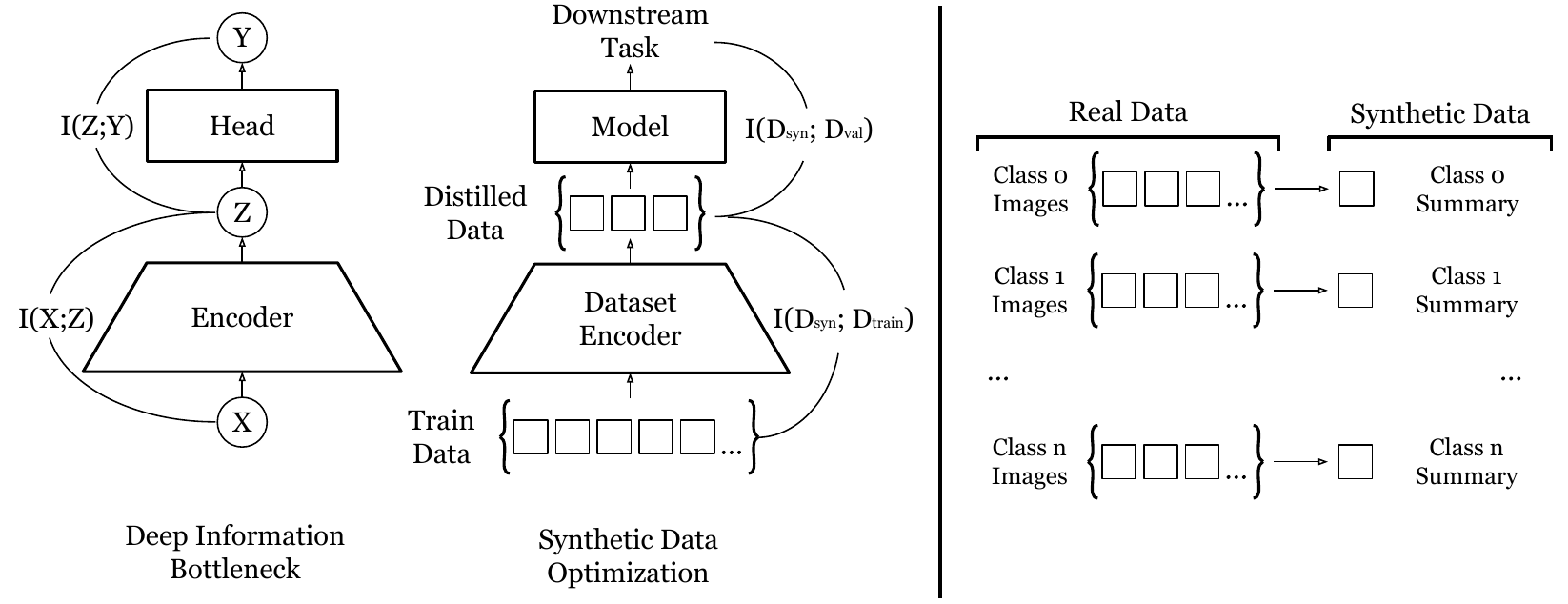}}
  \caption{On the left, we draw a parallel between intra-model representation learning with the deep information bottleneck. On the right, we show an example of using dataset distillation to compress a training set to 1 image per class for classification tasks.}
  \label{fig:datasetdistillation}
\end{figure}

To better study the information about the data summaries in this work, we employ the Information Bottleneck (IB) framework \cite{IB}. The IB framework aims to find a compressed representation $Z$ of some data $X$ while retaining as much information about a target variable $Y$ as possible. Overall, we extract relevant information from some signal about a target. The IB objective can be formally defined as:
\[\min_{p(z|x)} I(X;Z) - \beta I(Z;Y)\]
such that $I(X;Z)$ is the mutual information between $X$ and $Z$, $I(Z;Y)$ is the mutual information between $Z$ and $Y$, and $\beta$ is the trade off parameter for data compression. Minimizing $I(X;Z)$ improves the generalization as we are creating a simpler latent space \cite{FlatMinima}. Maximizing $I(Z;Y)$ preserves the relevant label information in the representation. It is important to note that $I(Z;Y) \leq I(X;Z)$ due to the Data Processing Inequality \cite{DataProcessingInequality}. This implies the representation cannot give more information about the target than what is in the input. With this in mind, we assume the following to make meaningful progress tied to the promising future of synthetic data.
\paragraph{The Synthetic Data Assumption:} \textit{For a downstream task and model $f$ trained on some real dataset $\mathcal{D}_{\text{real}}$ produced through traditional data collection techniques, there exists a synthetic dataset $\mathcal{D}_{\text{syn}}$ with the same spatial dimension that trains $f$ to achieve similar or better performance on the downstream task in non-trivially fewer datapoints than $\mathcal{D}_{\text{real}}$.}

Real datasets, for example image datasets of real-life objects, are limited by the physics of how they are obtained. Synthetic data has no such bounds: the set of all physically-possible images could be considered as a subset of all possible synthetic images. Synthetic data has the capacity to be much more feature-dense than physically collected data. Similar to how we consider intra-model representation learning in Figure \ref{fig:datasetdistillation} with the Deep Information Bottleneck \cite{DeepIB}, we instead aim to focus on better representations of the input before it reaches a model. 



This work aims to provide a foundation for future studies as more optimal representations of data will continue to be discovered by researchers, on models and tasks where this assumption holds. We explore this emerging paradigm by studying the effects of synthetic data using dataset distillation on structural learning methods such as neural network pruning. Therefore, we generalize framework to create some synthetic dataset $\mathcal{D}_{\text{syn}}$ such that we maximize the performance of a model trained on it. 
\begin{equation}
    \min_{\mathcal{D}_{\text{syn}}} L(\Phi(\mathcal{D}_{\text{syn}}); \mathcal{D}_{\text{val}}) \text{ s.t. } |\mathcal{D}_{\text{syn}}| < |\mathcal{D}_{\text{train}}|
\end{equation}
The loss here could be any evaluation metric on a downstream task. For example in image classification, you could optimize for low error on the validation set given some model with parameters $\Phi(\mathcal{D}_{\text{syn}})$. As shown on the left in Figure \ref{fig:datasetdistillation}, we can view synthetic data under this framework to better understand the dynamics of information flow. Instead of viewing $X$ is seen as a single input and $Z$ as the representation of that input, we aim to generate a compressed set of inputs $\mathcal{D}_{\text{real}}$ into a smaller set of the same spatial dimension, $\mathcal{D}_{\text{syn}}$. We want to maximize the useful label information of this representation set. So, we follow a similar Lagrangian:
\begin{equation}
    \min_{p(\mathcal{D}_{\text{syn}}|\mathcal{D}_{\text{train}})} I(\mathcal{D}_{\text{train}};\mathcal{D}_{\text{syn}}) - \beta I(\mathcal{D}_{\text{syn}};\mathcal{D}_{\text{val}})
\end{equation}
This formulation shows how we are compressing the irrelevant, input-specific information in the training set, yet maintaining relevant information to perform well on the downstream task. While $p(z|x)$ can be optimized through the encoder in the original formulation, we instead optimize the synthetic dataset encoder / generator instead. An optimal synthetic dataset, $\mathcal{D}_{\text{syn}}^*$ would inherently be more generalizable than $\mathcal{D}_{\text{train}}$ on downstream tasks as it contains all the relevant information with less complexity. With this, we expect a model trained on this data to require less complexity as well.

\section{Challenges with Distilled Data and Label Information}

Current dataset distillation methods do not properly maintain relevant information on the downstream task for harder datasets, but they do sufficiently minimize the image-specific information resulting in over-compression. These issues arise as a result of the high computational complexity of optimizing a synthetic training set. Current methods cannot maintain performance as the dataset or model complexity increases, limited to convolutional networks on toy datasets like CIFAR and MNIST. For example, we use Efficient Dataset Condensation \cite{EfficientDatasetCondensation} in this paper which achieves 50.6, 67.5, or 74.5\% accuracy on CIFAR-10 depending the the size of the synthetic dataset (1,10, or 50 images per class or ipc) or 72.8 \& 76.6\% accuracy on ImageNet-10 \cite{ILSVRC15}. While the aim of dataset distillation literature is to have a small synthetic set to improve training efficiency on other architectures, less aggressive compression may be needed. We suggest that efficiency is not the only use case or goal with distilled data. Compressing the training set in theory should result in better generalization according to the IB framework, as long as relevant information is maintained.

\section{Performance of Distilled Pruning}
\begin{figure}[htb]
\centering
  \includegraphics[width=12cm]{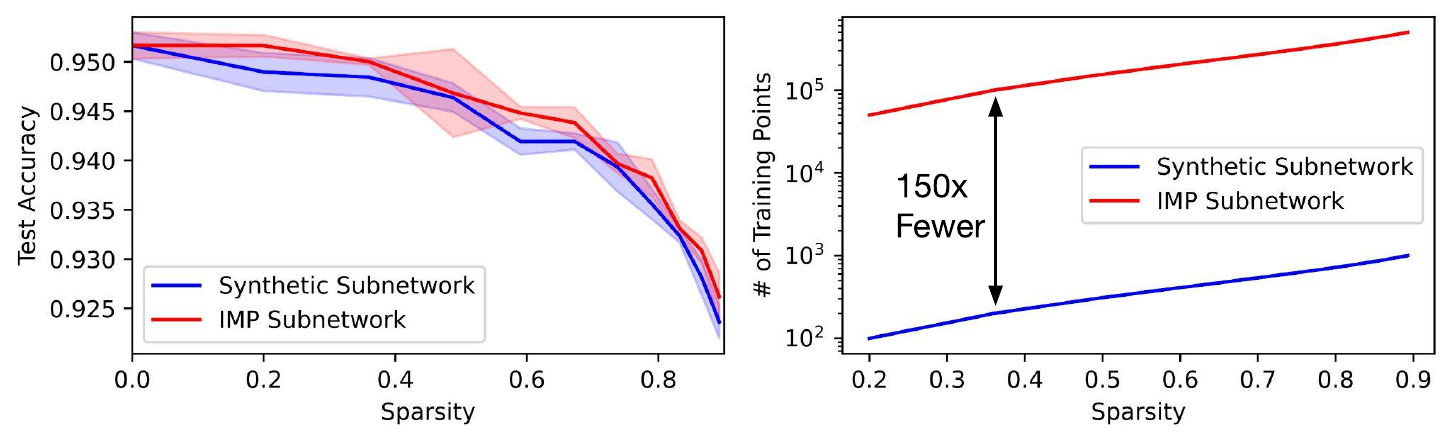}
  \caption{Performance of Distilled Pruning vs Traditional IMP on ResNet-18 \& CIFAR-10. The distilled dataset consisted of 10 images per class. Error bars are plotted as we average across 4 seeds. The plot on the right measures the amount of data points used in training to find a sparsity mask at x sparsity. Note that in IMP we are not finding "lottery tickets" since we rewind back to initialization for both methods, not to an early point in training.}
  \label{fig:SVATTM}
\end{figure}
We see in Figure \ref{fig:SVATTM} that lower test accuracy after training on distilled data does not completely translate in neural network pruning. Despite only achieving 72.8\% after training on distilled CIFAR-10 with 10 ipc from IDC \cite{EfficientDatasetCondensation}, we can use this low performance to choose weights to prune matching the performance on IMP on smaller datasets. To be clear, distilled data is only used to find a sparsity mask, the sparse model is ultimately trained on real data for validation. As previously mentioned, this performance does not hold as dataset complexity increases, we find that CIFAR-100 can perform decently at lower sparsities, but does not handle extreme sparsities ($>90\%$) with larger models like ResNet-18. In those cases, the distilled data on CIFAR-100 is not maintaining task-relevant information for the model. Since distilled data contains less outliers, and is a largely more "generalizable" dataset, we find that the sparsity mask pruned weights that control the fine grained details of the real data.

\section{Overview of Synthetic Subnetworks}
Previous experiments using dataset distillation compressed the original training set down to 10 images per class (IPC). We examine how varying the compression rate of dataset distillation affects the performance and stability of synthetic subnetworks. Across datasets, performance of synthetic subnetworks will be compared to its IMP counterpart. Therefore, we plot performance as the ratio of synthetic subnetwork accuracy over IMP subnetwork accuracy. We compare stability similarly against the IMP subnetworks. We plot stability as the barrier height ratio between the Synthetic and IMP subnetworks. Barrier height is measured as the distance between the trained loss and the loss of the interpolated model from our linear mode connectivity experiments. Note, on ImageNet-10 we exhibit a negative barrier height at high sparsity.

\begin{figure}[htb]
  \centering
  \makebox[0pt]{\includegraphics[width=5.5in]{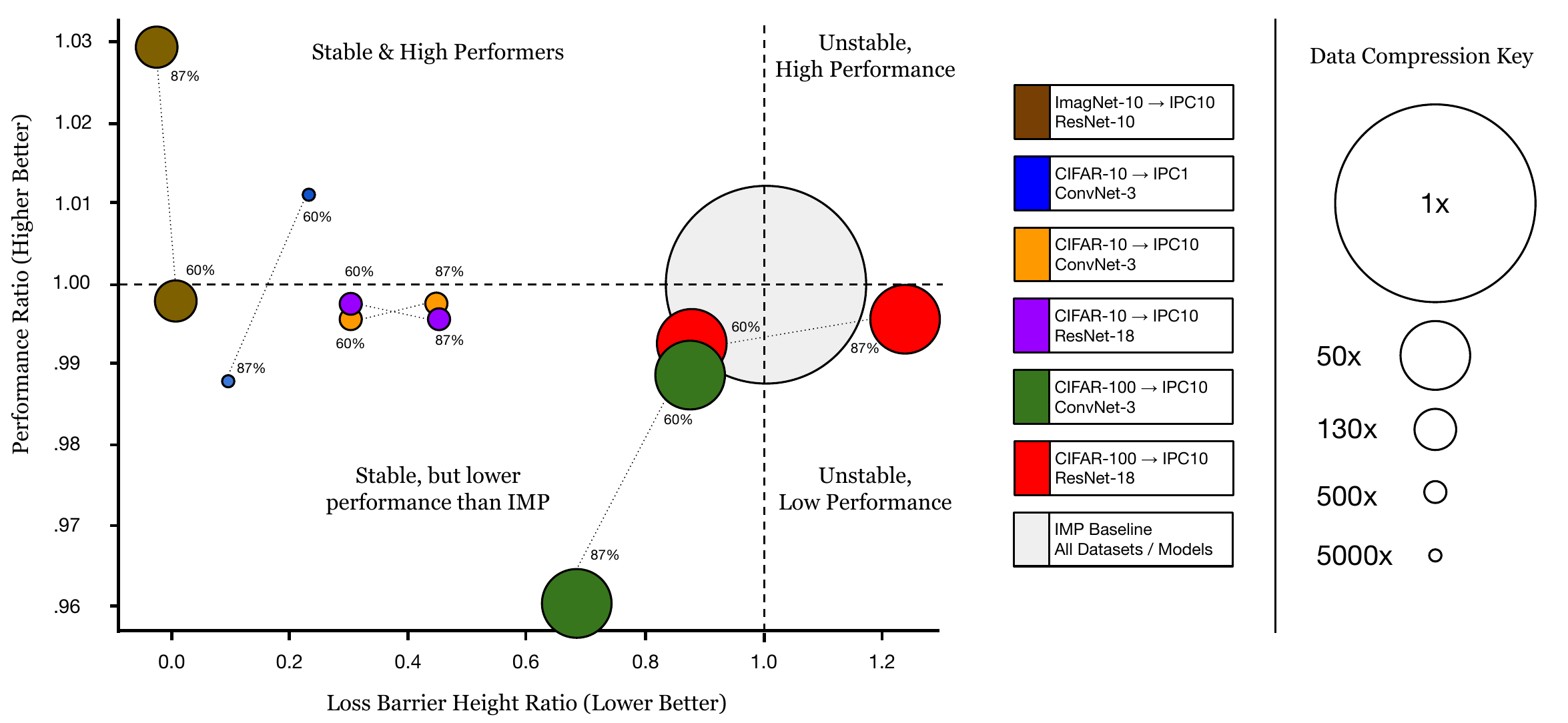}}
  \caption{Comparison of synthetic subnetworks and IMP subnetworks across models \& datasets. Both the 60\% and 87\% sparsity model is chosen for each example and connected with a dotted line. These are compared to IMP's 60\% and 87\% sparsity respectively. The size of the marker represents the amount of data compression through dataset distillation used to find the sparsity mask. For example, distilling CIFAR-10 (default 5,000 images per class) down to 10 images per class is a 500x compression ratio. Performance is measured as the ratio of test accuracy between Synthetic and IMP subnetworks. Loss Barrier Height is measured by subtracting the trained loss and the halfway interpolated trained loss from linear interpolation experiments. }
  \label{fig:scatterplot}
\end{figure}

\section{Approximating the Loss Landscape Sharpness}
For a more quantitative approach to measuring sharpness of the loss landscape, we measure the diagonal of the hessian for a batch.

\begin{center}
\begin{tabular}{||c | c | c | c | c ||} 
 \hline
 Subnetwork & Sparsity & Min/Max & Mean $\pm$ Std & Avg Magnitude \\ [0.5ex] 
 \hline\hline
 Dense & 0 & -1.847 / 1.344 & $.0048 \pm .035$ & .0067\\ 
 \hline
 IMP & 60\% & -3.073  / 4.510  & $.0051 \pm .0559$ & .0087\\
 \hline
 IMP & 90\% & -6.323 / 6.127  & $.0035 \pm .0594$ & .0060\\
 \hline
 Synthetic & 60\% & -0.982 / 2.093 & $.0047 \pm .0286$ & .0061 \\
 \hline
 Synthetic & 90\% & -2.338 / 1.730  & $.0037 \pm .0354$ & .0053\\ [1ex] 
 \hline
\end{tabular}
\end{center}

We see that across every measure of the distribution of the diagonal hessian, small amounts of pruning in traditional IMP lead to sharper landscapes. This smooths out as we approach higher sparsities though. With our synthetic subnetworks, the distribution is much closer to zero with less variance. We see the same pattern of smoother values as sparsity increases. While smoothness from increased sparsity may be a result of lower dimensionality of the landscape, synthetic subnetworks are still behaving radically different than IMP-chosen subnetworks at the same sparsity which can also be seen in the Figure \ref{fig:loss_landscape} visualization.

\end{document}